%% file: TaxonomyValidation.tex
\pdfoutput=1

\documentclass[letterpaper, 10 pt, conference]{ieeeconf}  

\IEEEoverridecommandlockouts                              

\usepackage{inputenc}
\usepackage{hyperref}
\usepackage{amsmath}
\usepackage{graphicx}

\usepackage{afterpage}
\usepackage{color}
\usepackage{xcolor,colortbl}
\usepackage{tikz}
\usetikzlibrary{arrows}

\title{\LARGE \bf
Analyzing Whole-Body Pose Transitions in Multi-Contact Motions}

\author{Christian Mandery, J\'ulia~Borr\`as, Mirjam J\"ochner and Tamim~Asfour     
\thanks{The research leading to these results has received funding from the European Union Seventh Framework Programme under grant agreement no 611832 (WALK-MAN) and grant agreement no 611909 (KoroiBot).}%
\thanks{The authors are  with the Institute for Anthropomatics and Robotics, Karlsruhe Institute of Technology, Germany. \newline
        {\tt\small \{mandery, julia.borrassol, asfour\}@kit.edu}}%
}

\begin{document}

\maketitle
\thispagestyle{empty}
\pagestyle{empty}


\begin{abstract}

When executing whole-body motions, humans are able to use a large variety of support poses which not only utilize the feet, but also hands, knees and elbows to enhance stability. While there are many works analyzing the transitions involved in walking, very few works analyze human motion where more complex supports occur.

In this work, we analyze complex support pose transitions in human motion involving locomotion and manipulation tasks (loco-manipulation). We have applied a method for the detection of human support contacts from motion capture data to a large-scale dataset of loco-manipulation motions involving multi-contact supports, providing a semantic representation of them. Our results provide a statistical analysis of the used support poses, their transitions and the time spent in each of them. In addition, our data partially validates our taxonomy of whole-body support poses presented in our previous work.

We believe that this work extends our understanding of human motion for humanoids, with a long-term objective of developing methods for autonomous multi-contact motion planning.

\end{abstract}


\section{Introduction}\label{sec:Introduction}
\input{sections/Introduction}

\section{Related Work}\label{sec:RelatedWork}
\input{sections/RelatedWork}

\section{Detection of Whole-Body Poses and Segmentation}\label{sec:segmentation}
\input{sections/Segmentation}

\section{Results}\label{sec:results}

\subsection{Statistical Analysis of the Detected Poses and Their Transitions}\label{subsec:statistics}
\input{sections/Results}

\subsection{Data-Driven Generation of a Transition Graph of Whole-Body Poses}\label{subsec:ExampleMotionAnalysis}
\input{sections/Taxonomy}

\section{Conclusions and Future Work}\label{sec:Conclusions}
\input{sections/Conclusions}



\bibliographystyle{IEEEtran}
\bibliography{TaxonomyValidation}

\end{document}

%% file: sections/Introduction.tex
While efficient solutions have been found for walking in different scenarios \cite{kajita_biped_2003,englsberger2013three}, 
including rough terrain and going up/down stairs, humanoid robots are still not able to robustly use their arms to gain stability, robustness and safety while executing locomotion tasks.

Robotics has approached this problem from a computational point of view (\cite{sentis_2010_compliant_TRO,lemerle2008robust, escande2010fast,lengagne_2013_generation,bouyarmane2011fem}). However, due to the complexity of the problem, these methods are still not completely successful. In this work, we propose to take a step back to analyze human motion in order to gain understanding of the processes humans make when using multi-contacts. 

Robotics in general, but particularly humanoid robotics, has always been inspired by biological human experience and the anatomy of the human body. However, human motions involving support contacts have almost not been studied \cite{nori2014whole} and even less how healthy subjects choose to make use of contacts with support surfaces. Works like \cite{johannsen2012contrasting} show that in a standing posture, reaching for a support contact provides augmented sensory information, reducing sway even if it is just through a "light touch". This shows that the ability of reaching for supports can be crucial to increase robustness in tasks that require balance like walking or general locomotion, but also for increasing maneuverability in complex manipulation tasks. Nevertheless, to execute such tasks in an autonomous way, we need to better understand the principles of whole-body coordination in humans, the variety of supporting whole-body postures available and how to transition between them.

\begin{figure}%
\includegraphics[width=\columnwidth]{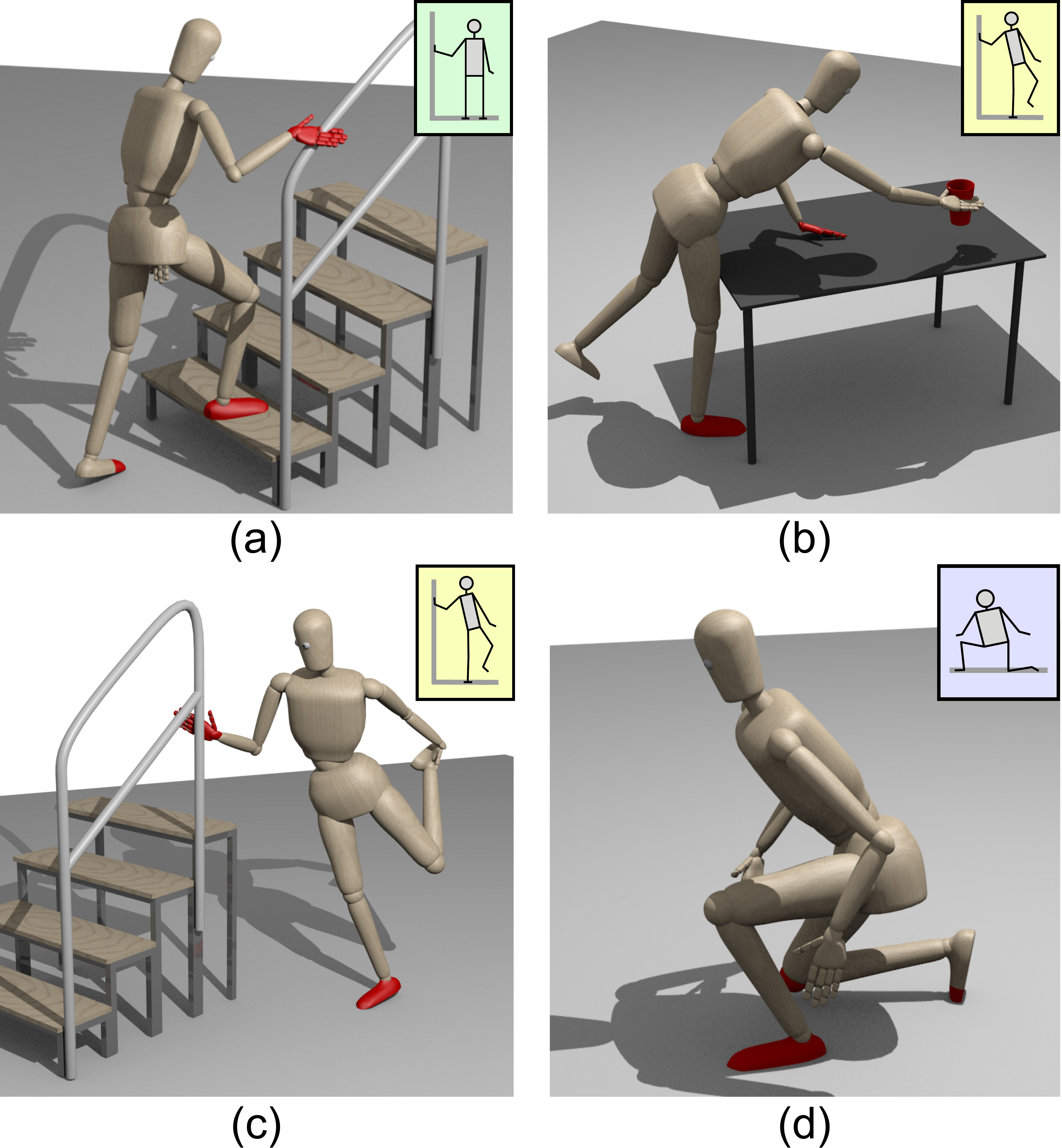}%
\caption{When performing locomotion (a), manipulation (b), balancing (c) or kneeling (d) tasks, the human body can use a great variety of support poses to enhance its stability. Automatically detecting such support contacts allows for an automatic identification of the visited support poses and their transitions.}
\label{fig:summary}
\end{figure}

In this work, we analyze real human motion data captured with a marker-based motion capture system and post-processed using our unifying Master Motor Map (MMM) framework \cite{Terlemez2014_MMM, Mandery2015}, to gain information about the poses that are used while executing
different locomotion and manipulation tasks like those shown in \autoref{fig:summary}. The analysis presented allows us to quantify the amount of time spent in each pose, classify transitions depending on their duration, and build a graph of pose transitions that can enlighten the difficult problem of finding motions that utilize multi-contacts to balance.

This paper is organized as follows. \autoref{sec:RelatedWork} briefly reviews related works. In \autoref{sec:segmentation}, we introduce our methodology to detect support poses. In \autoref{sec:results}, we apply our method to a large set of motions and analyze the resulting data. Finally, \autoref{sec:Conclusions} summarizes our contributions and gives prospects of future work.

%% file: sections/RelatedWork.tex
There are very few works studying how we transition between different support poses when performing movements. Contributions analyzing human data focus on specific postures for specific tasks, such as how to optimally hold on a handrail on a moving vehicle \cite{sarraf2014maintaining} or use hand support to better resist perturbations \cite{babivc2014effects}.
However, other areas of robotics like grasping have greatly benefited from the study of human hand poses. Three decades ago, they started analyzing human data to simplify the space of possible grasps \cite{cutkosky_taxonomy_1989,iberall1997human}. Works like \cite{ciocarlie2007dimensionality,santello1998postural} show that, although the hand posture space is highly dimensional, the majority of useful grasps can be described by a small number of discrete points in this space. Grasp taxonomies have been very relevant and successful \cite{cutkosky_taxonomy_1989, Kamakura_taxonomy_1989, feix_taxonomy_2009}, providing a wide variety of applications in grasp synthesis and autonomous grasp planning. Only a few works have tried to extend these concepts to whole-body motion \cite{forner2005principal}. In our recent work \cite{taxonomyJulia}, we have proposed a taxonomy of whole-body poses that use the environment to balance  based on a combinatorial approach of all the possible contacts using humanoid limbs. The current work provides a partial validation of the proposed taxonomy using real human data.

From the robotics community, there has been significant interest in improving balance control procedures beyond the double foot support \cite{sentis_2010_compliant_TRO,lemerle2013humanoid} with efficient path planners \cite{bouyarmane2009potential,lengagne_2013_generation} that need to solve computationally costly optimization problems under constraints \cite{escande2010fast}. However, these solutions are still not optimal, as planners are either very computationally costly \cite{lengagne_2013_generation,bouyarmane2009potential,bouyarmane2011fem} or only locally optimal and not applicable in an autonomous way \cite{sentis_2010_compliant_TRO,lemerle2008robust, escande2010fast}. Solutions to deal with the autonomous decision making \cite{philippsen2009bridging,yoshida2005humanoid,salini2011synthesis} provide interesting approaches but still do not scale to complex scenes. Each of these layers of the problem is a demanding problem on its own and the connection between all of them remains as one of the future challenges in robotics.

Our work relates to this literature in the long-term goal, but we want to approach the problem from a different point of view. Relying on human motion provides us with many transition motions that can be transferred to robots \cite{do2008imitation, Naksuk2005HumanToHumanoidMotionTransfer, Matsui2005generating} and stored as, {\em e.g.}, dynamic movement primitives (DMP) \cite{ijspeert2013dynamical}. There have been many works on DMPs for the whole body, showing that they can be adapted to different situations and sequenced \cite{Ude2010a, Kulic2012IncrementalLearning, kulvicius2012joining}. Other works do motion synthesis  \cite{arikan2003motion, yamane2011human}, usually based on different segmentation techniques. There has been extensive work on these segmentation techniques for human motion \cite{Kulic2009OnlineSegmentation, meier2011movement, jenkins2002deriving, janus2005unsupervised, Waechter2015}. 

Existing works to detect support contacts use video data \cite{yu2006detection}, tracking algorithms to estimate ground reaction forces \cite{brubaker2009estimating}, markers attached to the shoes to detect only floor contacts \cite{karvcnik2003using} or minimal oriented bounding boxes to detect links in contact without assuming environmental knowledge \cite{lengagne2012retrieving}.

%% file: sections/Segmentation.tex
\begin{table*}[bth]
 \caption{Evaluation of error of segmentation method}
\label{tab:evaluation}
\begin{center}

 \begin{tabular}{|l|c|c|c|c|l|} \hline 
{\bf Description}	     & {\bf \# Motions}&  {\bf Av. \# Poses}   & {\bf Av. \# Incorrect}& {\bf Av. \# Missed}	& {\bf Notes} \\ \hline
 \multicolumn{6}{|l|}{ {\bf Locomotion tasks}} \\ \hline
downstairs w. handle							& 10 		&	12.3		&	0.1	    &	2.5		& m: d.f.s. i: lost hand support	\\ \hline
upstairs w. handle								& 19		&	17.05		&	0.26	&	0.16	& m: d.f.s.	i: lost hand supports \\ \hline
upstairs, turn and downstairs			& 7		&	29.714		&	0.143	&	4	&	m: d.f.s. i: lost hand support \\ \hline
walks w. hand sup. to avoid obst.	& 5 		&	13.2		&	0.4	&	2.2	&	 m: d.f.s. i: lost hand support\\ \hline
walk over beam w. handle					& 5		&	19.4		&	0.2	&	0	&		\\ \hline
\multicolumn{6}{|l|}{ {\bf Loco-Manipulation tasks}} \\ \hline
kick box with foot w. hand sup.		& 6		&	12.5		&	0.33	&	1.167 	& m: d.f.s. i: lost hand support\\ \hline
lean to place a cup on table 			&	6		&	15.33		& 0.17  &   0    	&i: incorrect foot support\\ \hline
lean to pick a cup on table				& 5		&	5				&	0			&	  0			&		\\ \hline
lean to pick a cup in air 				& 7		& 15 			& 0.14 	&		0			&i: lost hand support \\ \hline
lean to wipe 											& 6		&	12.5		& 0.5		&   0			&m: d.f.s. at start\\ \hline
bimanual pick and place 					& 6		& 	13.833		&	0.833	&	0.667	&	 m: d.f.s. i: lost foot support \\ \hline
pick up from floor w. hand sup.		& 3  		&	4.67	 &	0.67 &	0	&	  i: extra hand support .	\\ \hline
\multicolumn{6}{|l|}{ {\bf Balancing tasks }} \\ \hline
push rec. fr. behind push w. lean	& 5  		&	6.2			&	0			&	0		&		\\ \hline
push rec. fr. left push w. lean		& 9		&	9.3			&	0.11	&	0		&		\\ \hline
inspect show sole w. sup.					& 2			&	11			&	0			&	0 	&\\ \hline
rec. fr. lost balance on 1 leg		& 5			&	10.8		&	0			&	0.2	&\\ \hline
lean on table w. hands						& 4			&	16.25		&	1.25	&	0		&	i: lost hand support	\\ \hline
\multicolumn{6}{|l|}{ {\bf Kneeling tasks }} \\ \hline
kneel down			& 4 		&	8				&	0	&	0	&		\\ \hline
kneel up				& 7			&	7.857		&	0	&	0	&		\\ \hline
\hline
{\bf Totals} 				& 121	& 	239.94 	&  	5.11 	& 	10.89			& \\ \hline
{\bf Percentages}		& 			&						&	2.13\%   	&	  4.53\%		& \\
\hline
 \end{tabular}
\end{center}
\text{Abbreviations: av. = average, w. = with, sup. = support, obst. = obstacle, fr. = from, rec. = recovery, i: incorrect, m: missed, d.f.s. = double foot support}
\end{table*}

\subsection{Motion Acquisition}

\begin{figure}
\centering
\includegraphics[width=0.48\columnwidth]{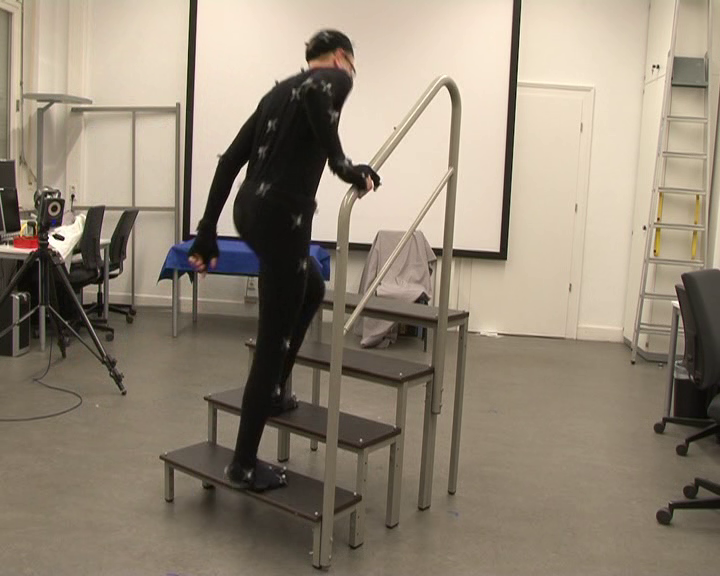}
\caption{Setup used for motion capture.}
\label{fig:mocap_setup}
\end{figure}

We captured 121 human motions using an optical marker-based Vicon MX motion capture system with 10 cameras.
A total of 56 passive reflective markers were attached to the human subject at characteristic anatomical landmarks. Subject was asked to perform different whole-body motions, described in \autoref{tab:percentagesTransitions}.
\autoref{fig:mocap_setup} illustrates the setup used for motion capture.
Details about the procedures used for motion acquisition, {\em e.g.} the marker set, can be found in \cite{Mandery2015} and online\footnote{\url{https://motion-database.humanoids.kit.edu/marker_set/}}.
After recording, the human motions were normalized and post-processed as described in \autoref{subsec:motion_processing}.

In addition to the human motion, we also captured the position and movement of objects and environmental elements.
To allow the reconstruction of object trajectories, a minimum number of three additional markers were placed on each object in a non-collinear manner.
Using manually created object models, object trajectories can then be estimated from the marker trajectories which allow the analysis of interaction between the human subject and these environmental entities \cite{Waechter2015}.
The KIT Whole-Body Human Motion Database \cite{Mandery2015} contains a large set of motions using this approach, providing raw motion capture data, corresponding time-synchronized video recordings and processed motions.

\subsection{Motion Processing}
\label{subsec:motion_processing}

The Master Motor Map (MMM) \cite{Azad2007_MMM, Terlemez2014_MMM} provides an open-source framework for capturing, representing and processing human motion.
It includes a unifying reference model of the human body for the capturing and analysis of motion from different human subjects.
The kinematic properties of this MMM reference model are based on existing biomechanical analysis by Winter \cite{winter2009} and allow the representation of whole-body motions using 104 degrees of freedom (DoF): 6 for the root pose, 52 for the body torso, extremities and head, and $2\cdot 23$ for the hands. For the analysis in this work, we have excluded the hand joints.

To be able to extract semantic knowledge from the recorded motions, we first need to transfer these motions to the MMM reference model, {\em i.e.} reconstruct joint angle trajectories from the motion capture marker trajectories in Cartesian space.
For this purpose, for every motion capture marker on the human subject, we place one corresponding virtual marker on the reference model.

Let ${\bf U}=({\bf u}_1, ..., {\bf u}_n)$ be an observation of the 3D positions of the $n$ captured markers and ${\bf x}=(p_x, p_y, p_z, \alpha, \beta, \gamma, \theta_1, ..., \theta_m)$ the vector describing the pose of the reference model, consisting of the root position and rotation of the model and its $m$ joint angle values. 
Additionally, let ${\bf V}({\bf x})=({\bf v}_1({\bf x}), ..., {\bf v}_n({\bf x}))$ be the positions of corresponding virtual markers as determined by the forward kinematics of the model.
The problem of determining the pose of the MMM reference model for a given marker observation ${\bf U}$ is then solved by minimizing
\begin{equation*}
f({\bf x}) = \sum\limits_i ({\bf u}_i - {\bf v}_i({\bf x}))^2
\end{equation*}
while maintaining the box constraints for $\theta_1, ..., \theta_m$ given by the joint limits of the reference model.
For every motion frame, this optimization problem is solved by using the reimplementation of the Subplex algorithm \cite{Rowan1990Subplex} provided by the NLopt library \cite{JohnsonNlopt} for nonlinear optimization.
Poses of objects involved in a motion are reconstructed from object markers in a similar way by using a joint-less six-dimensional pose vector.

\subsection{Extraction of Whole-Body Poses}

Support poses of the human subject are detected by analyzing the relation of the MMM reference model to the floor and environmental elements.
For this purpose, we only consider objects which exhibit low movement during the recorded motion as suitable environmental elements to provide support.
For every motion frame, we use the forward kinematics of the reference model to calculate the poses of the model segments that we consider for providing supports.
These model segments represent the hands, feet, elbows and knees of the human body.

A segment $s$ of the reference model is recognized as a support if two criteria are fulfilled.
First, the distance of $s$ to an environmental element must be lower than a threshold $\delta_{dist}(s)$.
Distances to environmental elements are computed as the distances between pairs of closest points from the respective models with triangle-level accuracy using Simox \cite{Vahrenkamp12b_Simox}.
Additionally, the speed of segment $s$, computed from smoothed velocity vectors, has to stay below a threshold $\delta_{vel}(s)$ for a certain number of frames, starting with the frame where the support is first recognized.
The thresholds are chosen empirically:
$\delta_{vel}=200\frac{mm}{s}$, $\delta_{dist}(Feet)=\delta_{dist}(Hands)=15mm$, $\delta_{dist}(Knees)=35mm$ and $\delta_{dist}(Elbows)=30mm$.

\begin{table*}[bt]
\caption{Percentages of appearances and time spent for each transition (\%appearance, \%time)}
\label{tab:percentagesTransitions}
\def\arraystretch{1.5}
\begin{tabular}{l|c|c|c|c|c|c||c|}
           & 1Foot & 1Foot-1Hand  &  2Feet&  2Feet-1Hand &   2Feet-2Hands & 1Foot-2Hands &   Totals x pose \\ \hline
1Foot&   			4.38\%, 5.69\%			& 9.30\%, 7.90\% 	&   22.90\%,   25.56\%	& 0.15\%, 0.26\%		&   --	& 	0.08\%, 0.04\%		& 36.81\%, 39.44\%\\ \hline
1Foot-1Hand & 9.15\%, 13.64\% &  1.81\%, 2.26\% &  0.08\%, 0.03\%		& 12.24\%, 16.59\%	&   0.08\%, 0.02\%	&    0.15\%,    0.02\%	& 23.51\%, 32.57\%\\ \hline
2Feet&      16.02\%, 10.05\%     & 0.15\%, 0.04\% &   $\times$	&  3.48\%, 2.23\%		& 0.08\%, 0.06\%		&   --  & 19.73\%, 12.38\% \\ \hline
2Feet-1Hand  & 0.23\%, 0.07\%		& 11.72\%, 4.38\%		&   4.61\%, 5.31\%	& $\times$	& 0.98\%, 0.15\%	& --	&   17.54\%, 9.92\%\\ \hline
2Feet-2Hands & --	& --	&  --		& 0.83\%, 1.22\%&   $\times$ 	& 0.68\%, 0.75\%		& 1.51\%, 1.97\%\\ \hline
1Foot-2Hands & -- &   0.53\%, 1.27\%	&  --	&   -- &   0.38\%, 2.45\%	& $\times$&  0.91\%, 3.72\%\\ \hline
\end{tabular}
\end{table*}

The support pose is defined by the contacts that are providing support to the subject.
We ignore parts of the motion where the human body is not supported at all as an empty support pose, {\em e.g.} during running.
Also, some practical assumptions are used, such as that a knee support also implies a foot support.

The video attachment shows some of the motions that were part of our evaluation along with detected support contacts and the resulting support poses. We have manually validated the segmentation method error by exploring frame by frame the detected support segments, showing the results in \autoref{tab:evaluation}. They show that about 4.5\% of the poses are missed, but the missed poses are always double foot supports (with or without hand). Only 2.1\% of the poses are incorrectly detected.

%% file: sections/Results.tex
\begin{figure*}[t]
\centering
\begin{tabular}{p{0.5\linewidth} p{0.5\linewidth}}
(a)\includegraphics[width= 0.42 \textwidth ]{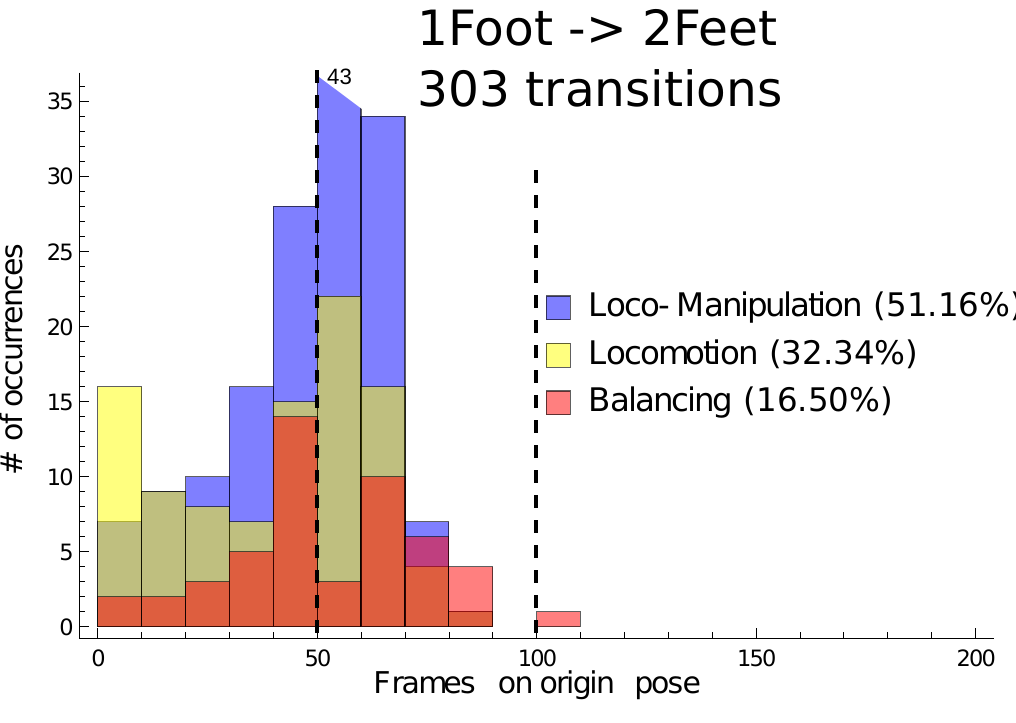} & (b)\includegraphics[width= 0.42 \textwidth ]{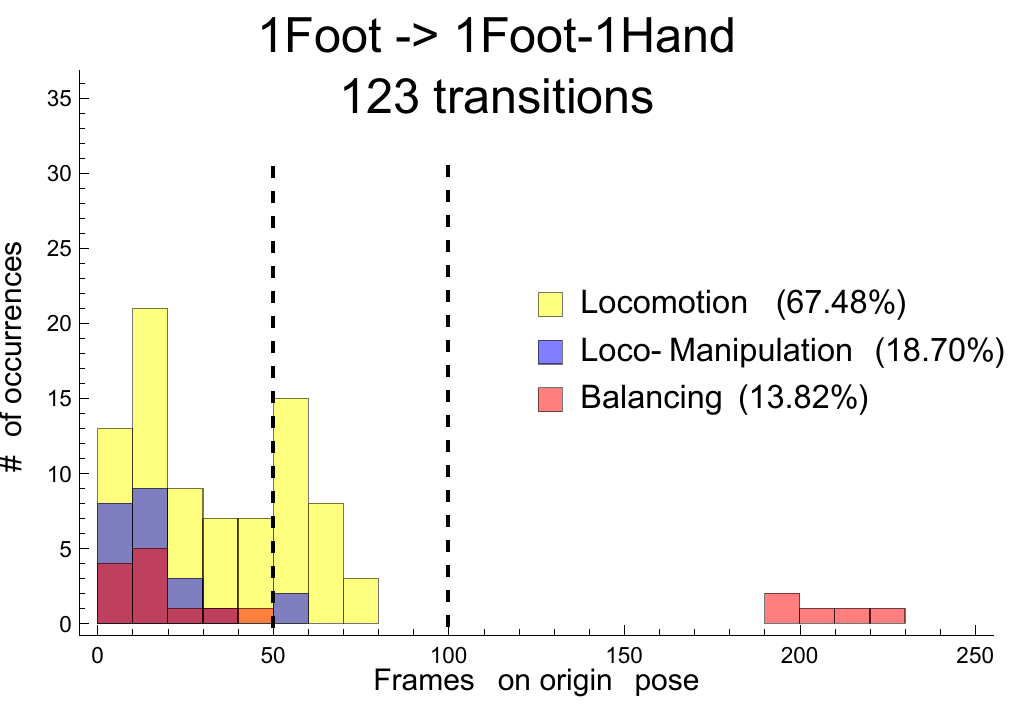} \\
(c) \includegraphics[width= 0.42 \textwidth ]{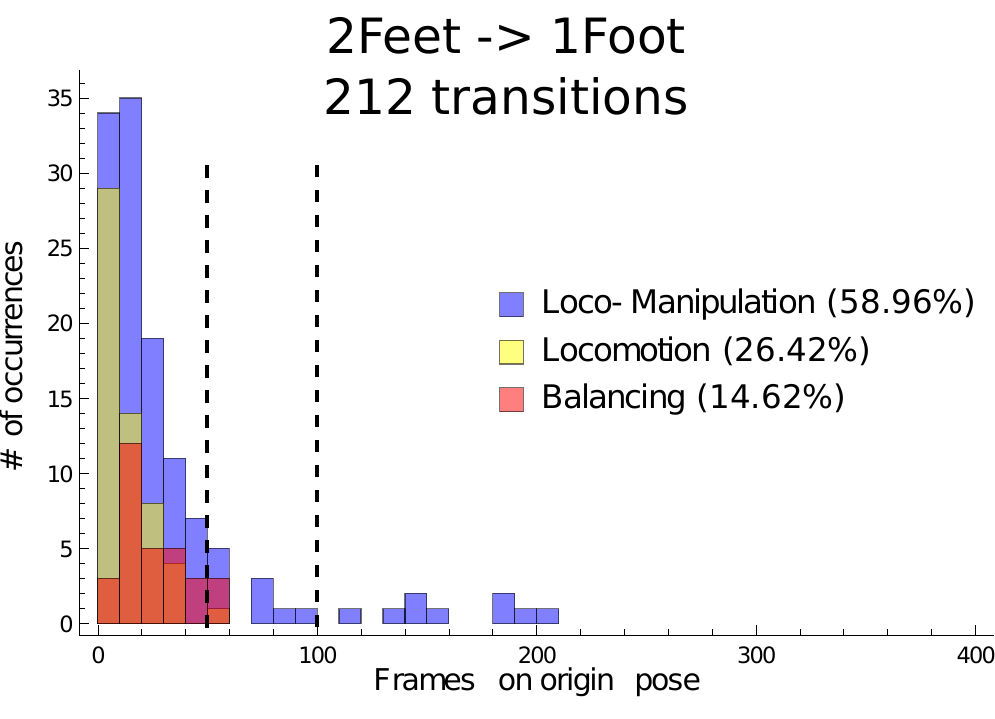} & (d)\includegraphics[width= 0.42 \textwidth ]{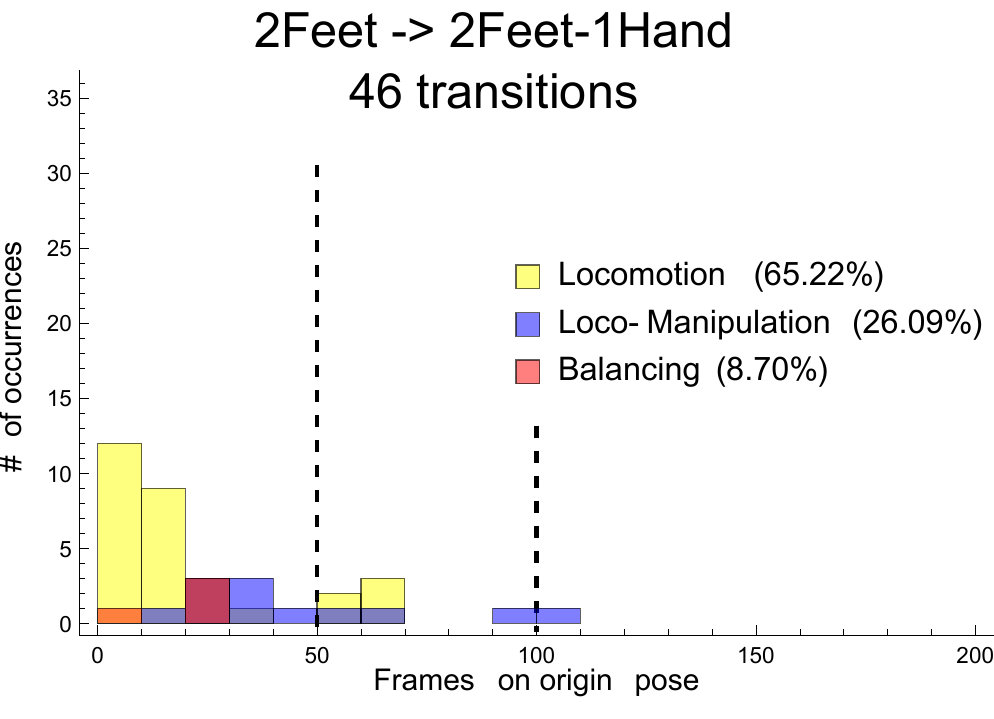} \\
(e)\includegraphics[width= 0.42 \textwidth ]{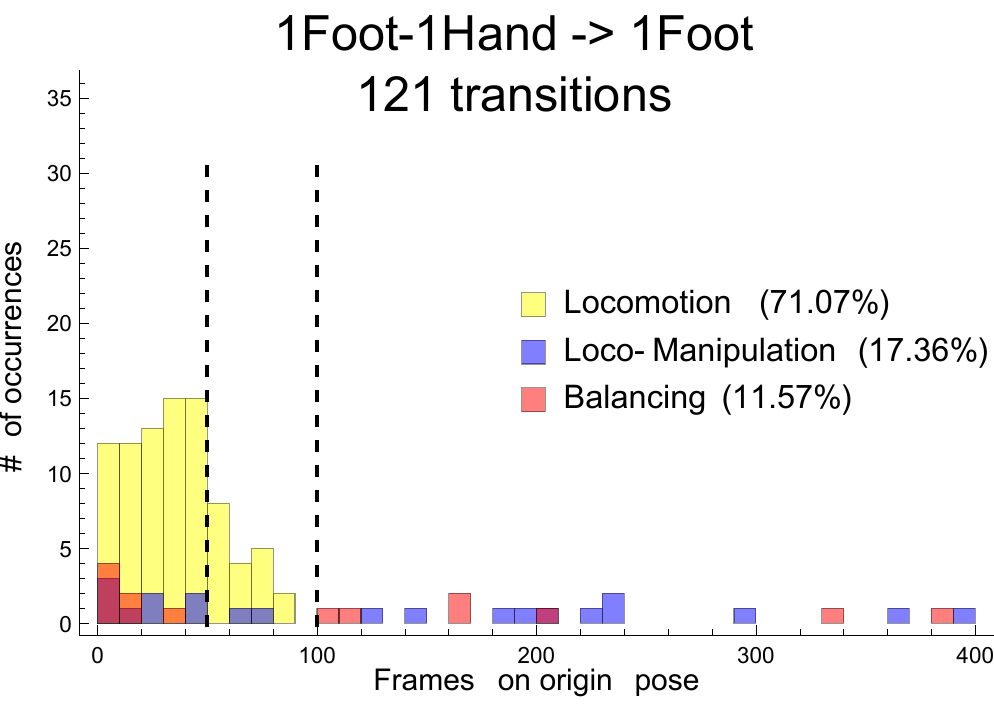} & (f)\includegraphics[width= 0.42 \textwidth ]{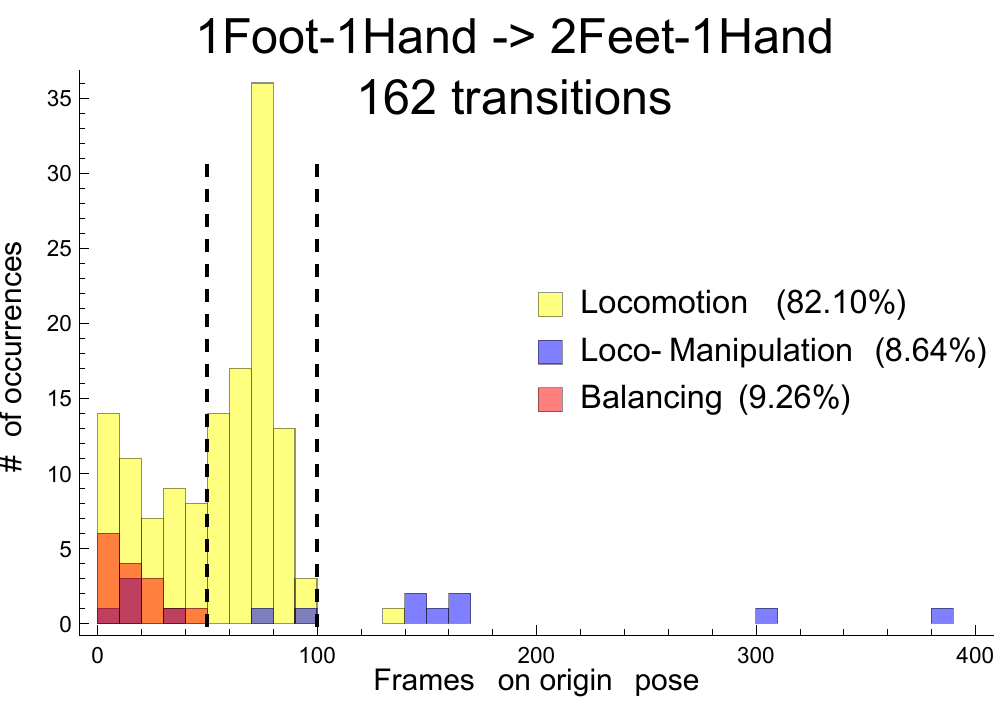} \\
(g)\includegraphics[width= 0.42 \textwidth ]{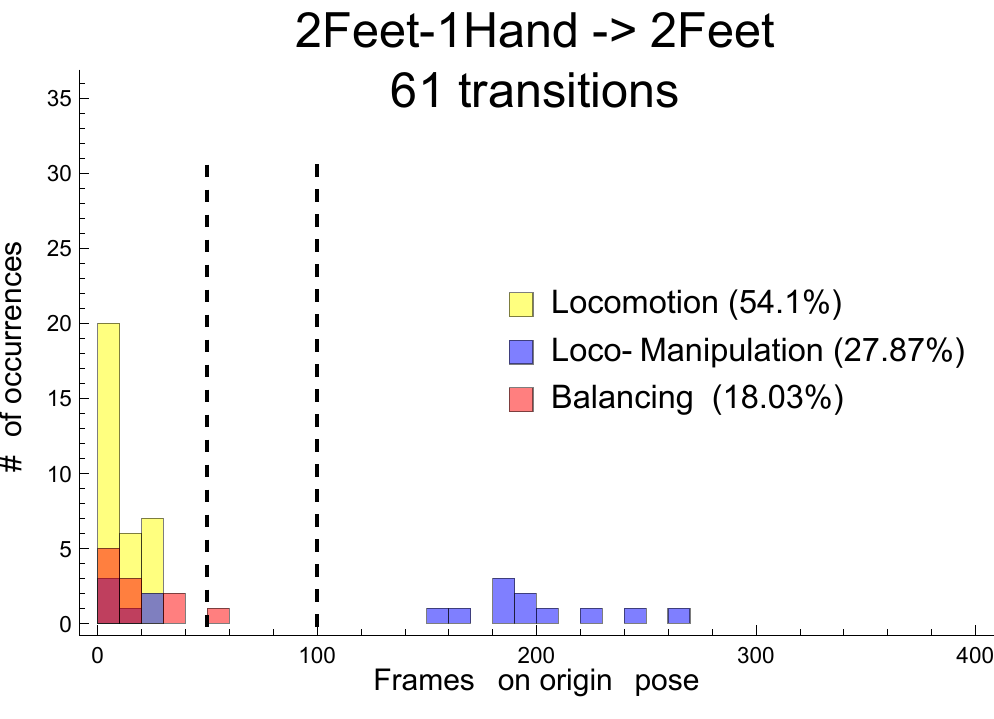} & (h)\includegraphics[width= 0.42 \textwidth ]{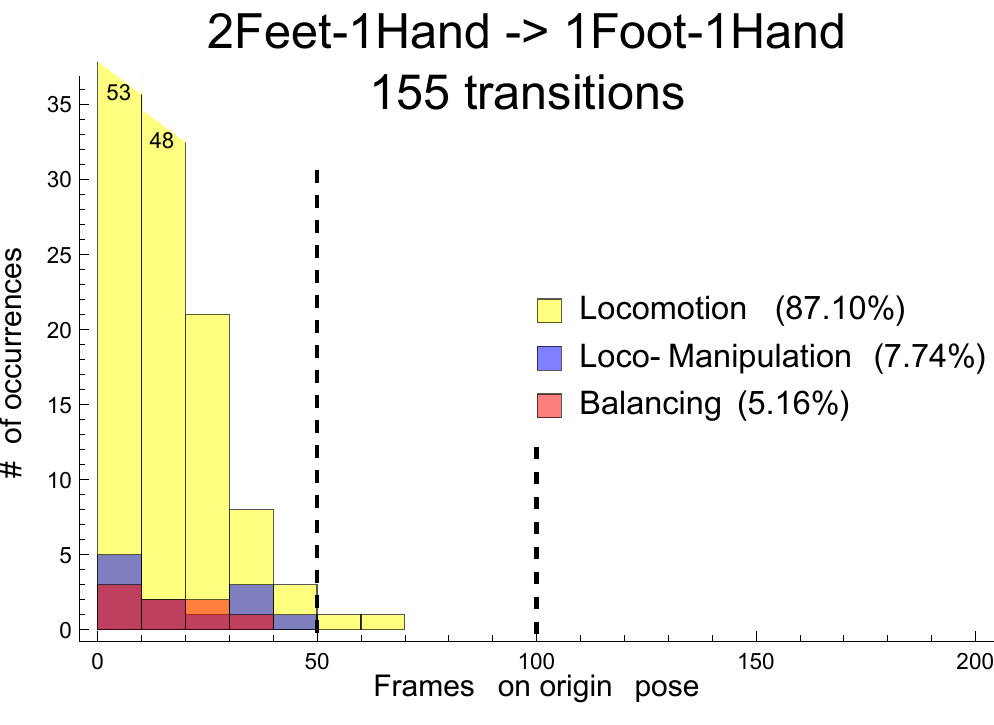} \\ 
\end{tabular}
\caption{Histograms showing the occurrences of frames spent in each transition. Bins are of 10 frames each. Dashed lines at 50 frames and 100 frames to clarify the different $x$ axis scales.}
\label{fig:histogram}
\end{figure*}

Without taking into account kneeling motions, we have recorded and analyzed 110 motions including locomotion, loco-manipulation and balancing tasks listed in \autoref{tab:evaluation}. In this section, we present some analysis on the most common pose transitions and the time spent on them. We ignore kneeling motions because we do not have enough data yet to get significant results.
In every motion, both the initial and the final pose are double foot supports and the time spent on these poses is arbitrary. Therefore, they have been ignored for the statistical analysis. Without counting them, we have automatically identified a total of 1323 pose transitions lasting a total time of 541.48 seconds (9.02 min). In \autoref{tab:percentagesTransitions}, each cell represents the transition going from the pose indicated by the row name to the pose indicted by the column name. In each cell, we show first the percentage of occurrence of the transition with respect to the total number of transitions detected, and secondly the percentage of time spent on the origin pose before reaching the destination pose, with respect to the total time of all motions. The last column is the accumulation of percentages per each pose, and the rows are sorted from the most to the least common pose.

It must be noted that the loop transitions 1Foot$\rightarrow$1Foot, and 1Foot-1Hand$\rightarrow$1Foot-1Hand are mostly missed double foot supports and we will not include them in the analysis.

According to \autoref{tab:percentagesTransitions}, the most common transitions are 1Foot$\rightarrow$2Feet (22.90\% of appearance) and 2Feet$\rightarrow$1Foot (16.02\% of appearance). These are the same transitions of walking that have been widely studied. Winter reported in \cite{winter1984kinematic} that depending on slow or fast walking, the interval of the time spent on 2Feet$\rightarrow$1Foot (double foot support) is 11--19 frames\footnote{All times are measured in frames, with motions recorded at 100 FPS.}, while for 1Foot$\rightarrow$2Feet (single foot support) it is 38--52 frames. Although all our motions contain some steps of normal walking, they also involve hand supports, and therefore, these transitions may show different time behaviors if they are part of a more complex set of transitions. 
We are interested in observing similar long and short locomotion transitions, but involving other poses. In addition, we find a third type of transition that usually lasts longer because it is supporting a manipulation task. The transitions in \autoref{tab:percentagesTransitions} where the time spent is proportionally larger than their frequency can give us the intuition that they may be either long locomotion transitions or support for manipulation tasks.

\begin{figure*}
\centering
\includegraphics[width=0.75\linewidth]{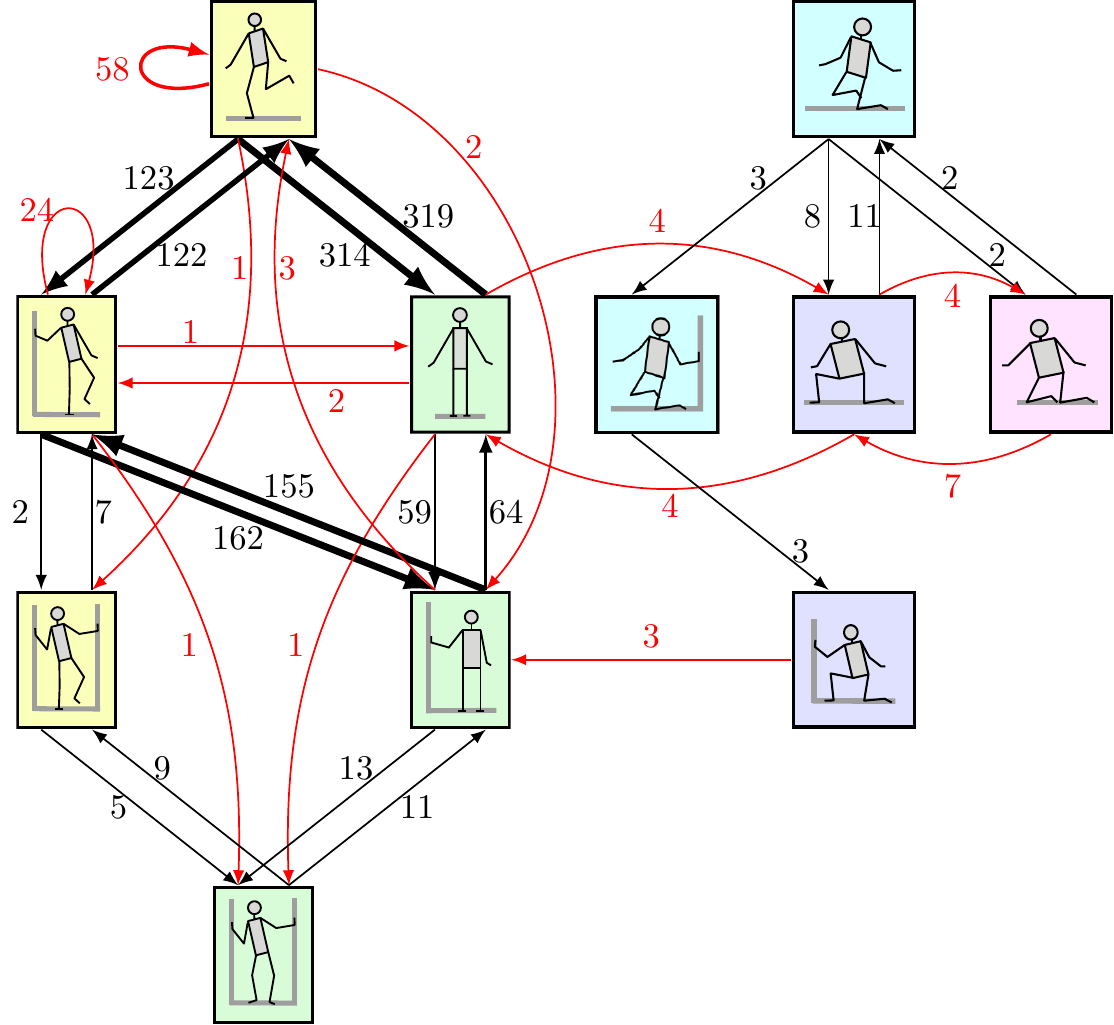}
\caption{Transition graph of whole-body pose transitions automatically generated from the analyzed motions. Labels on edges indicate the number of transitions found of each type.}
\label{fig:taxonomy}
\end{figure*}

\subsection{Analysis of the Time Spent per Transition}
To study the time spent in each transition in more detail, \autoref{fig:histogram} shows the histograms of time spent in the most common pose transitions. In yellow, we show all transitions involved in locomotion tasks, and we can observe that the histograms in (a), (b) and (f) show bimodal distributions. For the first two cases, we could fit a mixture distribution of 2 normals, with parameters $N(\mu=11.76, \sigma=8.60)$ and $N(\mu=53.89, \sigma=11.35)$ for the plot (a) and $N(\mu=15.9905, \sigma=9.6776)$ and $N(\mu=55.8507, \sigma=8.7216)$ for the plot (b), with a confidence probability of 0.969 and 0.944 respectively. This indicates that the 1Foot$\rightarrow$2Feet transition can play the role of a long locomotion transition with mean 53 frames, but can also be a short transition with times around 11 frames, and similarly for (b). For the plot (f), we could fit the two normals $N(\mu=36.47, \sigma=26.59)$ and $N(\mu=73.11, \sigma=8.319)$, but with only 0.80 of confidence. We need more data to verify these mean values. Still, inspecting the histogram it is clear that the transition 1Foot-1Hand$\rightarrow$2Feet-1Hand can act as a long transition with mean times of around 70 frames Other transitions like (c), (d), (g) and (h) are clearly short transitions. Plot (c) corresponds to 2Feet$\rightarrow$1Foot (the usual walking double foot support) that for locomotion tasks is clearly on the short duration, with 76.8\% of the cases below 20 frames (91\% below 30).

In blue, we show the loco-manipulation tasks. These tasks include walking, but also transitions for supporting the manipulation task. Note that transitions to support the manipulation are not very frequent because there is only one per motion, while transitions for walking are the majority, shown in plots (a) and (c). As expected, (a) shows long locomotion transition types, while (c) shows short ones. However, in (c) we see some long-lasting poses. Inspecting the data task by task, we see that these happen in the bimanual pick and place of the big box, because the double foot pose supports the action of crouching to pick up the box. In the remaining plots, we can see other transitions supporting manipulation. For instance, the ones in plots (e) and (f) correspond to some of the motions of leaning to wipe, reach or place, where the subject uses a 1Foot-1Hand to perform the task 
(as in \autoref{fig:summary}-(b)), while in other lean actions, the subject uses the 2Foot-1Hand pose, shown in plot (g).

Finally, balancing tasks are plotted in red. They consist mostly of very fast transitions because motions are very fast, especially after pushes. As before, plots (a) and (c) accumulate the poses of the walking parts of the motions. In (b), the 200 frames lasting transitions correspond to the balancing on one foot, lasting until the subject loses balance and needs to lean with the hand. After that, the 1Foot-1Hand pose is used until balance is recovered, showing as long transitions in plot (e). Other transitions supporting a task in plot (e) correspond to inspecting the shoe sole, that is supported by a 1Hand-1Foot pose. We do not show the histograms containing four supports because we do not have enough instances of them, but they all happen during balancing tasks.

The motions recorded for this work can be found in the KIT Whole-Body Human Motion Database \cite{Mandery2015}\footnote{See \url{https://motion-database.humanoids.kit.edu/details/motions/<ID>/} with ID $\in$ \{383, 385, 410, 412, 415, 456, 460, 463, 515, 516, 517, 520, 521, 523, 527, 529, 530, 531, 597, 598, 599, 600, 601, 604, 606, 607\}.}.

%% file: sections/Taxonomy.tex
\begin{figure*}
\centering
\includegraphics[width=\textwidth]{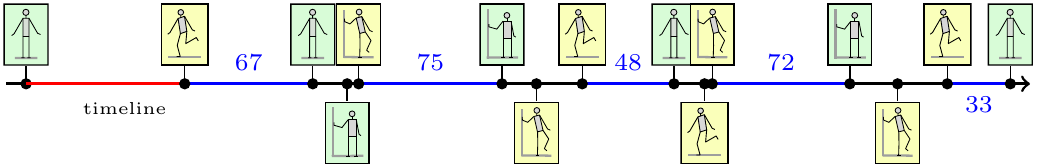}
\caption{Output of the segmentation for one of the motions upstairs with handle. The segment shown in red represents the initial pose transition, that has an arbitrary length. Blue segments represent transitions where the foot swings. Blue labels indicate transition durations. We can see that the human alternates between single foot support swing and 1Foot-1Hand support swing using the handle.}
\label{fig:timelineExample}
\end{figure*}

In \cite{taxonomyJulia}, we proposed a full taxonomy of whole-body support poses that was based on a combinatorial approach considering all the relevant contacts with the body. The current work has been inspired by our taxonomy, and one of its objectives is to validate the transitions that we proposed. However, we can only provide a partial validation, because our theoretical taxonomy contained poses with holds (like when hands grasp a handle) and also arm and forearm supports, that do not appear in any of the motions analyzed here. Also, we need more data on kneeling poses to reach more of the four support poses.

\autoref{fig:taxonomy} shows the automatically generated transition graph, considering also the start and end poses of each motion.
Each edge corresponds to a transition, and their labels to the number of times we have found it.
 Edges plotted in red correspond to transitions where two simultaneous changes of contacts occur. In our theoretical taxonomy \cite{taxonomyJulia}, we assumed that only one change of support should be allowed per transition. While this is still desirable for robotics, it is also obvious that some human transitions involve two contact changes. For instance, in push recovery motions, humans usually lean on the wall using both arms at the same time to increase security and robustness. Many of the red edge transitions in \autoref{fig:taxonomy} occur in balancing tasks.

In the transition graph shown in \autoref{fig:taxonomy}, we can quickly see that red edges are of significantly lower frequency than the black ones, except the loop edges in the 1Foot and 1Hand-1Foot poses, that are caused by either jumps or missed double foot supports. They correspond to the 4.5\% transitions missed by our segmentation method reported in \autoref{tab:evaluation}.

This data-driven transition graph is influenced by the type of motions we have analyzed, using only one handle or one hand support. Only balancing poses reach the four support poses. In future work, we will analyze walking motions with handles on both sides. 

\autoref{fig:timelineExample} shows the timeline of a motion where the subject goes upstairs using a handle on his right side. In blue, we show the long locomotion transitions. The supporting pose for these transitions alternates between 1Foot-1Hand, used to swing forward the foot not in contact, and 1Foot, used to swing forward both the handle hand and the foot not in contact. This is because we only provide one handle. Another interesting thing to notice is that the short locomotion transitions appear in clusters, composed by a sequence of two transitions. We have observed this in many of the motions and we have observed that the order of the transitions inside these clusters does not matter, just the start and end poses. We believe that each cluster could be considered as a composite transition where several contact changes occur. As future work, we want to detect and model these clusters to identify rules that allow us to automatically generate sequences of feasible transitions according to extremities available for contacts.

%% file: sections/Conclusions.tex
We have presented an analysis of support poses of more than 100 motion recordings showing different locomotion and manipulation tasks. Our method allowed us to retrieve the sequence of used support poses and the time spent in each of them, providing segmented representations of multi-contact motions.

Although the most common pose transitions are the ones involved in walking, we have shown that the 1Foot-1Hand and the 2Foot-1Hand poses also play a crucial role in multi-contacts motions. We have classified our data into short and long locomotion transitions and transitions for supporting a task, depending on the time spent on them. We have observed that very short locomotion transitions are found in clusters that can be grouped as complex transitions with more than one contact change. The data-driven generated taxonomy validates the transitions proposed in our previous work. We believe that our
motions segmented by support poses and time spent per transition provides a meaningful semantic representation of a motion. 

This work opens the door to many exciting future directions. First, we are interested in analyzing our motion representations to find semantic rules that can help define new motions for different situations, with the objective of building a grammar of motion poses. Storing each transition as motion primitives, we are also interested in performing path planning at a semantic level based on support poses.

Finally, we are still assuming very simplified poses that do not consider directions of support, represented by simple sketch figures. 
However, for each class of poses there is an infinite number of possible body configurations depending on location and orientation of contacts. Future work directions include finding the most relevant {\em whole-body eigen-grasps}, that is, we will perform principal component analysis to reduce the dimensionality of the space that can realize each pose.

In conclusion, this work presents a step further in the comprehension of how humans can utilize their bodies to enhance stability for locomotion and manipulation tasks.